\documentclass[10pt,twocolumn,letterpaper]{article}

\usepackage[pagenumbers]{iccv} 
\usepackage{algorithmic}

\usepackage[ruled]{algorithm2e}
\usepackage{url}
\usepackage{booktabs}
\usepackage{graphicx}
\usepackage{booktabs} 
\usepackage{multirow}
\usepackage{bbm}
\usepackage{subcaption}

\definecolor{iccvblue}{rgb}{0.21,0.49,0.74}
\usepackage[pagebackref,breaklinks,colorlinks,allcolors=iccvblue]{hyperref}

\def\paperID{11090} 
\def\confName{ICCV}
\def\confYear{2025}

\title{When Preferences Diverge: Aligning Diffusion Models with Minority-Aware Adaptive DPO}

\author{
    Lingfan Zhang\textsuperscript{1}\thanks{Equal contribution.} \quad
    Chen Liu\textsuperscript{2}\footnotemark[1] \quad
    Chengming Xu\textsuperscript{3} \quad
    Kai Hu\textsuperscript{4} \quad
    Donghao Luo\textsuperscript{3} \quad
    Chengjie Wang\textsuperscript{3} \quad\\
    Yuan Yao\textsuperscript{2} \quad
    Yanwei Fu\textsuperscript{1}\thanks{Corresponding author.} \quad\\
    \textsuperscript{1}Fudan University \quad
    \textsuperscript{2}The Hong Kong University of Science and Technology \quad\\
    \textsuperscript{3}Tencent \quad
    \textsuperscript{4}Carnegie Mellon University
}

\begin{document}
\maketitle
\begin{abstract}
In recent years, the field of image generation has witnessed significant advancements, particularly in fine-tuning methods that align models with universal human preferences. This paper explores the critical role of preference data in the training process of diffusion models, particularly in the context of Diffusion-DPO and its subsequent adaptations. We investigate the complexities surrounding universal human preferences in image generation, highlighting the subjective nature of these preferences and the challenges posed by minority samples in preference datasets. Through pilot experiments, we demonstrate the existence of minority samples and their detrimental effects on model performance. We propose Adaptive-DPO -- a novel approach that incorporates a minority-instance-aware metric into the DPO objective. This metric, which includes intra-annotator confidence and inter-annotator stability, distinguishes between majority and minority samples. We introduce an Adaptive-DPO loss function which improves the DPO loss in two ways: enhancing the model's learning of majority labels while mitigating the negative impact of minority samples. Our experiments demonstrate that this method effectively handles both synthetic minority data and real-world preference data, paving the way for more effective training methodologies in image generation tasks.
\end{abstract}  
\section{Introduction}
\label{sec:intro}

Recent advances in diffusion models have revolutionized text-to-image generation, enabling unprecedented visual quality and diversity. These models have demonstrated remarkable capabilities in synthesizing highly detailed and realistic images from textual descriptions, making them invaluable tools for creative industries, education, and entertainment. However, the practical deployment of these models hinges on their ability to align with human preferences, ensuring that the generated outputs not only meet technical standards but also resonate with aesthetic and contextual expectations of users. To address this, researchers have adapted Reinforcement Learning from Human Feedback (RLHF)~\cite{ouyang2022training}, a technique originally developed for language models, to the diffusion paradigm. Methods like Diffusion-DPO~\cite{wallace2024diffusiondpo} bypass the need for an explicit reward model by directly optimizing preference alignment through pairwise comparisons, achieving state-of-the-art results in aligning diffusion models with human preferences.

\begin{figure}[t]
    \centering
    \includegraphics[width=\linewidth]{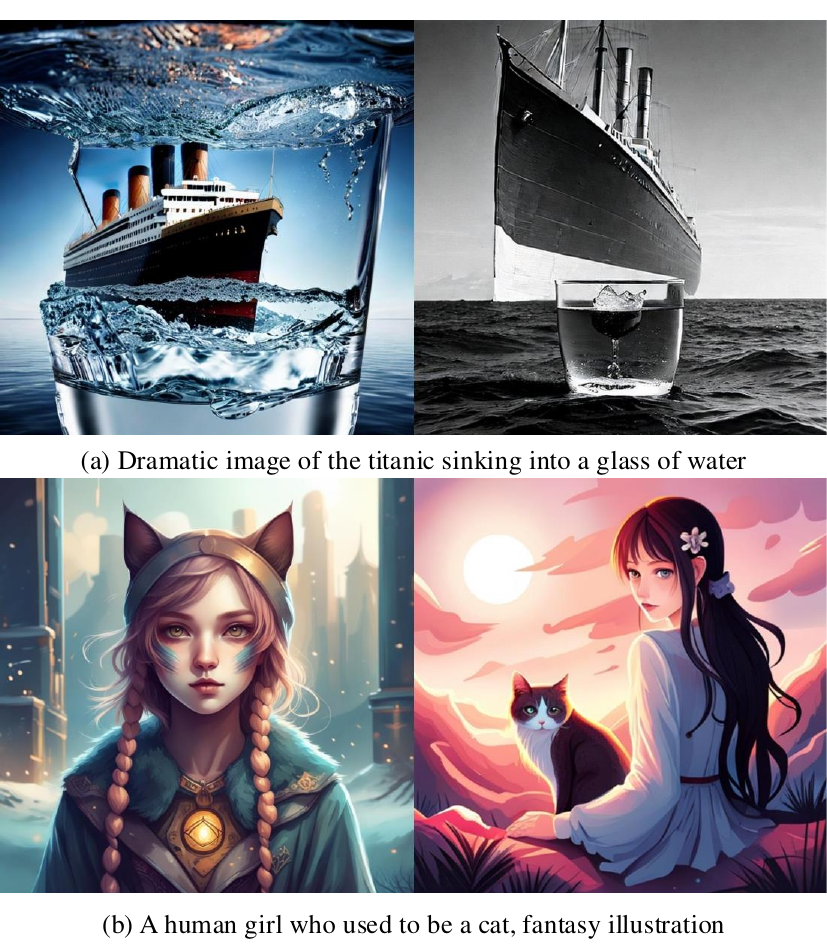}
    \caption{Two examples from Pick-a-Pic~\cite{kirstain2024pick}. For both rows, the winning images are the right ones.}
    \label{fig:teaser}
    \vspace{-0.2in}
\end{figure}

Despite these advancements, a critical assumption underlying these methods remains relatively less unexplored: the \textit{uniformity and reliability} of preference datasets. Current approaches implicitly assume that preference labels uniformly reflect universal human judgment. However, such a notion fails to account for the inherent subjectivity of visual aesthetics and the practical realities of crowd-sourced annotation. In fact, preference datasets exhibit a hidden dichotomy. The majority of annotations reflect consensus criteria such as image fidelity, prompt alignment, and overall visual appeal. These are the patterns that methods like DPO are expected to learn and optimize for. However, a significant minority of annotations can significantly deviate from this consensus. These minority data can be attributed to two primary sources. (1) \textbf{Erroneous annotations} arise from human errors, such as annotator laziness, misunderstandings, or misinterpretations of the task. For instance, in Fig.~\ref{fig:teaser}(a), the left image is obviously better in terms to either image quality and text fidelity, while the annotator labels the right one as the winning image. (2) \textbf{Subjective divergences} represent valid but niche preferences that conflict with majority standards. These include stylistic biases, cultural differences, or personal tastes that lead annotators to prefer outputs that deviate from the consensus. For example, we asked 25 participants with different genders and ages about which image in Fig.~\ref{fig:teaser}(b) is better, considering both image aesthetics and text-image alignment. Among them, 18 participants think the left one is better, while the others think the opposite, which demonstrates the subjectivity divergence.

In this paper, we advocate exploring the impact of such a dichotomy for DPO. Specifically, we first conduct a pilot study where we manually creates minority samples by randomly flipping the original preference. The experiments demonstrate that the presence of these minority samples poses a significant challenge for DPO. For example, flipping 20\% data degrades ImageReward of SD1.5 finetuned with DPO from 0.16 to 0.00, equivalent to 52\% of the performance gain achieved by DPO. 

To address these challenges, we propose Adaptive-DPO, a novel preference learning framework that autonomously identifies and suppresses problematic minority samples while preserving legitimate majority preferences. Our approach leverages two key innovations. First, we introduce a self-driven minority-aware metric that jointly models intra-annotator confidence and inter-annotator stability. Intra-annotator confidence measures the certainty of predictions by evaluating agreement across multiple model checkpoints, while inter-annotator stability quantifies the variance in predictions across different training stages. This dual mechanism allows our metric to distinguish between the major preference criteria and other minority preferences without relying on external reward models or costly relabeling efforts. Second, we propose modulating the preference learning process based on the designed metric. This includes instance-specific reweighting to suppress minority samples and adaptive margins to enhance supervision from majority samples. Together, these terms enable Adaptive-DPO to robustly align diffusion models with human preferences while mitigating the impact of subjective annotations. 

Extensive evaluations demonstrate the effectiveness of Adaptive-DPO across multiple benchmarks and model architectures. Concretely, we compare our method with previous methods using both SD1.5 and SDXL as backbones, on benchmarks including Pick-a-Pic~\cite{kirstain2024pick} and HPDv2~\cite{wu2023hps}. Our Adaptive-DPO significantly outperforms the competitors, showcasing its ability to handle ambiguous and subjective annotations. We further analyze our method with abundant ablation studies, showing that Adaptive-DPO is not only better than methods such as re-filtering the training data, and but also can generalize well to other preference optimization variants such as IPO~\cite{azar2024general}. In summary, the contributions of this work are as follows:

\begin{itemize}
    \item We provide the first systematic study of the majority/minority preference dichotomy in diffusion alignment, shedding light on a critical but understudied aspect of preference learning.
    \item We introduce the novel minority-instance-aware metric by considering both intra-annotator confidence and inter-annotator stability.
    \item We propose Adaptive-DPO, which can not only mitigate the negative impact of minority samples but also enable the model to better leverage reliable majority annotations.
    \item We conduct comprehensive empirical validation across multiple metrics, datasets, and backbones, demonstrating the generalizability and scalability of our approach.
\end{itemize}

\section{Preliminaries}

\subsection{RLHF and Reward Model}
Reinforcement Learning from Human Feedback (RLHF) fine-tunes large models by aligning them with human preferences through preference optimization with two phases: 

\begin{itemize}
    \item \textbf{Reward Modeling}: This phase employs the Bradley-Terry (BT) model to learn a reward function from pairwise data \((x_w, x_l, c)\), where \(x_w\) is the preferred answer and \(x_l\) is the less preferred one. The loss function is defined as:
\begin{equation}\label{eq:rlhf_rm}
   \mathcal{L}_R(r_{\phi}, \mathcal{D}) = -\mathbb{E}_{(c, x_w, x_l)}\left[\log \sigma(r_{\phi}(c, x_w) - r_{\phi}(c, x_l))\right]
\end{equation}
where $\sigma$ denotes sigmoid function, $r_{\phi}$ is the reward model parameterized by $\phi$.
    \item \textbf{RL Finetuning}: This phase uses the learned reward model to provide feedback to the language model, applying the Proximal Policy Optimization (PPO) algorithm:
\begin{equation}\label{eq:rlhf_rl}
   \max_{\pi_{\theta}} \mathbb{E}_{c\sim \mathcal{D}, x\sim \pi_{\theta}(x \mid c)}\left[r_{\phi}(c, x)\right] - \beta\mathbb{D}_{\textrm{KL}}\left[\pi_{\theta}(x\mid c) \| \pi_{ref}(x\mid c)\right]
\end{equation}
\end{itemize}

\subsection{DPO and Diffusion-DPO Objective}
DPO (Direct Preference Optimization) and Diffusion-DPO both utilize pairwise human preferences without requiring a reward model. 

For DPO, the explicit solution derived from the RLHF framework is:
\begin{equation}\label{eq:dpo_solution}
\pi_r(y\mid x) = \frac{1}{Z(c)}\pi_{ref}(x\mid c)\exp\left(\frac{1}{\beta}r(c, x)\right)
\end{equation}
where $Z(c) =\sum_{x}\pi_{ref}(x\mid c)\exp\left(\frac{1}{\beta}r(c, x)\right)$, this means:
 \begin{equation}\label{eq:main_eq}
    r(c,x) =\beta \log \frac{\pi_r(x\mid c)}{\pi_{ref}(x\mid c)} + \beta \log Z(c).
\end{equation}

This leads to the DPO optimization objective:
\begin{equation}
\begin{aligned}\label{eq:dpo}
& \mathcal{L}_\text{DPO}(\pi_{\theta}; \pi_{ref}) = -\mathbb{E}_{(x, y_w, y_l) \sim \mathcal{D}} \\
&\left[\log \sigma\left(\beta \log \frac{\pi_{\theta}(y_w\mid x)}{\pi_{ref}(y_w\mid x)} - \beta \log \frac{\pi_{\theta}(y_l\mid x)}{\pi_{ref}(y_l\mid x)}\right)\right]
\end{aligned}
\end{equation}
For Diffusion-DPO, the objective is formulated as:
\begin{equation}\label{eq:diffusion-dpo}
\begin{split}
     &\mathcal{L}(\theta)
    = - \mathbb{E}_{(x^{w}, x^{l}) \sim \mathcal{D}, t\sim \mathcal{U}(0,T), x^{w}_{t}\sim q(x^{w}_{t}|x^{w}), x^{l}_{t} \sim q(x^{l}_{t}|x^{l})} \\
    &\log\sigma \left(-\beta T \omega(\lambda_t) \left(
    \| \epsilon^w -\epsilon_\theta(x^{w}_{t},t)\|^2_2 - \|\epsilon^w - \epsilon_{\text{ref}}(x^{w}_{t},t)\|^2_2 \right. \right. \\
    &\left. \left. - \left( \| \epsilon^l -\epsilon_\theta(x^{l}_{t},t)\|^2_2 - \|\epsilon^l - \epsilon_{\text{ref}}(x^{l}_{t},t)\|^2_2\right)\right)\right)
\end{split}
\end{equation}
where \(\epsilon\) represents the noise prediction network and \(t\) is the denoising timestep.

\section{Pilot Study: Influence of Minority Samples \label{sec:pivot2}}

To verify if the minority preference samples can have a detrimental effect on the model's learning, we in this section conduct an intuitive pilot study. Concretely, since it is hard to detect the minority samples in real world data, we propose to flip a random part of the original preference labels, i.e. the winning images are turned into losing ones. To show such simulation is reasonable, we in Fig.~\ref{fig:pilot} provide an explanation, in which we perform the same flipping process to 1000 samples of which 10\% are labeled as minority samples. As can be found in the figure, as the proportion used for flipping getting larger, more original majority samples are changed into minority ones, indicating larger proportion of minority data. Formally, for both SD1.5 and SDXL, we randomly flip 10\%, 20\% and 30\% data, train them with original Diffusion-DPO, and record several metrics such as ImageReward (IR)~\cite{xu2024imagereward}, PickScore (PS)~\cite{kirstain2024pick}, Aesthetic Score (Aes)~\cite{schuhmann2022laion} and HPS~\cite{wu2023better}.

\begin{table}[htb]
\hfill
\parbox{.38\linewidth}{
\captionof{figure}{Proportion of majority/minority at different flip ratios.}
\label{fig:pilot}
\vspace{-0.1in}
\centering
\resizebox{\linewidth}{!}{%
    \includegraphics[width=0.5\linewidth]{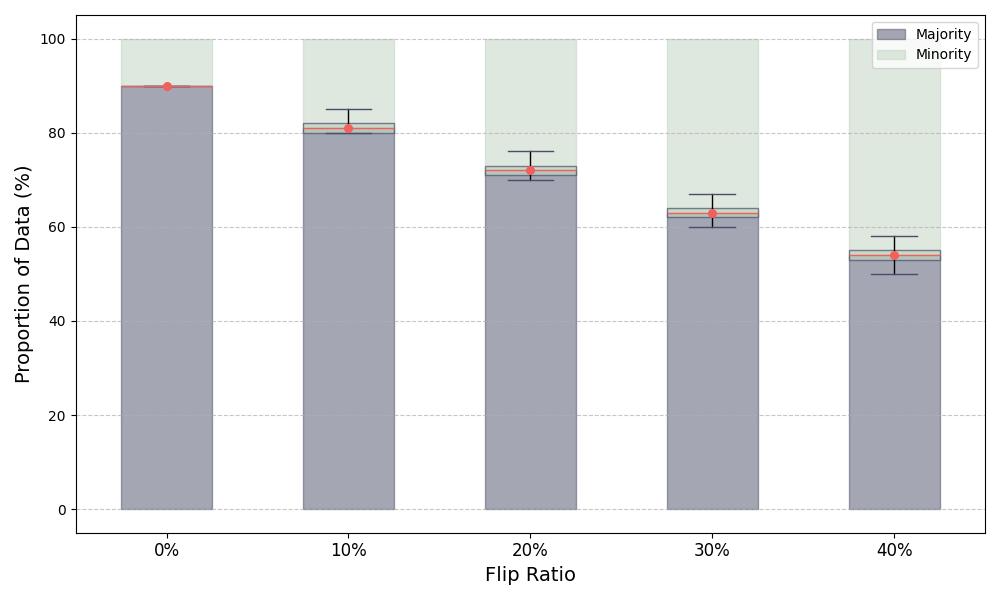}
}
}
\hfill
\parbox{.45\linewidth}{
\caption{DPO results with different noise level. Larger metric, better performance.}
\label{tab:pivot2}
\vspace{-0.1in}
\centering
\resizebox{\linewidth}{!}{%
\begin{tabular}{cccccc}
\toprule

Model & Flip rate (\%) & IR ($\uparrow$) & PS ($\uparrow$) & Aes ($\uparrow$) & HPS ($\uparrow$) \tabularnewline
    \midrule

    \multirow{4}{*}{SD1.5} & 0 & \textbf{0.16} & \textbf{21.05} & \textbf{5.31} & \textbf{26.43} \tabularnewline
    & 10 & 0.05 & 20.91 & 5.27 & 26.32 \tabularnewline
    & 20 & 0.00 & 20.83 & 5.24 & 26.24 \tabularnewline
    & 30 & -0.05 & 20.72 & 5.21 & 26.14 \tabularnewline
    \midrule
    \multirow{4}{*}{SDXL} & 0 & \textbf{0.87} & \textbf{22.52} & 5.89 & \textbf{27.32} \tabularnewline
    & 10 & 0.79 & 22.53 & \textbf{5.90} & 27.21 \tabularnewline
    & 20 & 0.70 & 22.38 & 5.86 & 27.09 \tabularnewline
    & 30 & 0.66 & 22.29 & 5.88 & 26.99 \tabularnewline

\bottomrule
\end{tabular}
}
}
\vspace{-0.15in}
\end{table}

As presented in Tab.~\ref{tab:pivot2}, we can find that with more minority samples, the fine-tuning process exhibits a significant decline in its efficacy, eventually resulting in a total deterioration of the model's performance. Specifically, for SD1.5, 30\% flipped samples can even lead to negative image reward. This indicates that the Diffusion-DPO algorithm can be easily affected by the minority samples contained in the training data. Moreover, it is noteworthy that our manually created minority samples are intrinsically different from the real-case minority samples. While our randomly selected minority data forms a uniform distribution, the distribution of real minority data could be related with many factors, such as the difficulty of annotation problems, the quality of data, etc. Therefore, it is important to design a new algorithm that is robust to both cases.

\section{Methodology}


To address the challenges posed by minority samples in preference datasets, we propose Adaptive-DPO, a novel framework designed to identify and suppress problematic minority samples while preserving the integrity of majority preferences. The key novlety of our approach include: (1) a self-driven metric that captures both intra-annotator confidence and inter-annotator stability, and (2) an Adaptive-DPO objective that prioritizes the learning of majority labels while mitigating the influence of minority samples.

\subsection{Measuring Minority Preferences}

In the training process of DPO, the supervision signal is derived from preference labeling.
As discussed in Sec.\ref{sec:intro}, the labeling process for preference data is not always reliable. Minority labels from two distinct sources can both undermine the efficacy of DPO and degrade model performance. Therefore, a robust minority-aware metric must account for both types of minority samples simultaneously. To this end, we begin by presenting a simple observation of the DPO formulation. Specifically, Eq.\ref{eq:dpo} can be reformulated as:
\begin{align}\label{eq:rewrite-dpo}
    \mathcal{L}_\text{DPO}(\pi_{\theta}; \pi_{ref}) &= -\mathbb{E}_{(x, y_w, y_l)\sim \mathcal{D}}\left[\log \sigma \left(\beta (\eta_{\theta}-\eta_{ref})\right)\right] \\
    \eta_{\theta} &= \log \frac{\pi_{\theta}(x^{w}|c)}{\pi_{\theta}(x^{l}|c)} \\
    \eta_{ref} &= \log\frac{\pi_{ref}(x^{w}|c)}{\pi_{ref}(x^{l}|c)}
\end{align}

For Diffusion-DPO, as formulated in Eq.\ref{eq:diffusion-dpo}, a similar structure can be derived. By minimizing $\mathcal{L}\text{DPO}$, the model is effectively guided to maximize $\eta{\theta} - \eta_{\text{ref}}$. This mechanism is analogous to optimizing with a binary hinge loss. Consequently, the DPO fine-tuning process implicitly steers the model to act as a binary preference classifier. Previous works\cite{liu2020early} have shown that such models tend to first learn correctly labeled samples, which constitute the majority of the training data, before addressing incorrectly labeled samples, which are often minority cases.
Inspired by prior methods in semi-supervised learning~\cite{cascante2021curriculum} and robust learning~\cite{goel2022pars}, we propose leveraging the predictions of fine-tuned models to instantiate a minority-aware metric. This metric consists of two main components, as described below.

\noindent\textbf{Intra-annotator confidence.} Samples with obvious preference patterns but incorrectly judged often exhibit the exact opposite preference criteria compared to the majority samples. For instance, images with lower quality or poorer text fidelity may be favored in some incorrectly judged pairs. Since the model can easily learn their features from the majority labels, the predictions for these samples will significantly differ from their original labels. As a result, the gap between the model's predictions (referred to as the bias) and the given labels remains substantial throughout the training process.
To formalize this, we first define a metric to measure the difference between the fine-tuned model and the reference model.

In detail, given a sample of pair, we have
\begin{equation}
\label{ltheta}
    \ell_{\theta} = \eta_{\theta}-\eta_{ref}
\end{equation}
The larger value of $\ell_{\theta}$ indicates a higher confidence.

Afterwards, we begin to define the detailed metric.
Suppose we have $M$ different models, which are instantiated as a main model finetuned by DPO and the historical EMA footprint of it, we give the definition for intra-annotator confidence as follows, 
\begin{equation}\label{eq:confidence_dpo}
    c_\theta(x) = 1-\frac{1}{M}\sum_{m=1}^{M} \sigma(\ell_{\theta^{(m)}}(x)\cdot\rho
  )
\end{equation}
For $c_{\theta}$, the large value indicates a large bias.

\noindent\textbf{Inter-annotator stability.} On the other hand, as for the samples with more complicated preference criteria and less obvious preference patterns, including them into training data can confuse the model and lead to more difficulty for the model to understand the target preference criteria. For such samples,  model's judgment may fluctuate because it cannot learn the information well, which would lead to quite unstable predictions regarding these pairs. Therefore, we use an inter-annotator stability term $s_\theta(x)$ to instantiate the \textit{variance} phenomenon. The detailed form is as follows, 
\begin{equation}\label{eq:std_dpo}
    s_\theta(x) =\frac{1}{M-1}\sum_{m=1}^{M} \big(\ell_{\theta^{(m)}}(x)-\frac{1}{M}\sum_{m=1}^{M} \ell_{\theta^{(m)}}(x)\big)^2
\end{equation}

\begin{figure}[t]
\begin{center}
\vspace{-0.2in}
\includegraphics[width=0.85\linewidth]{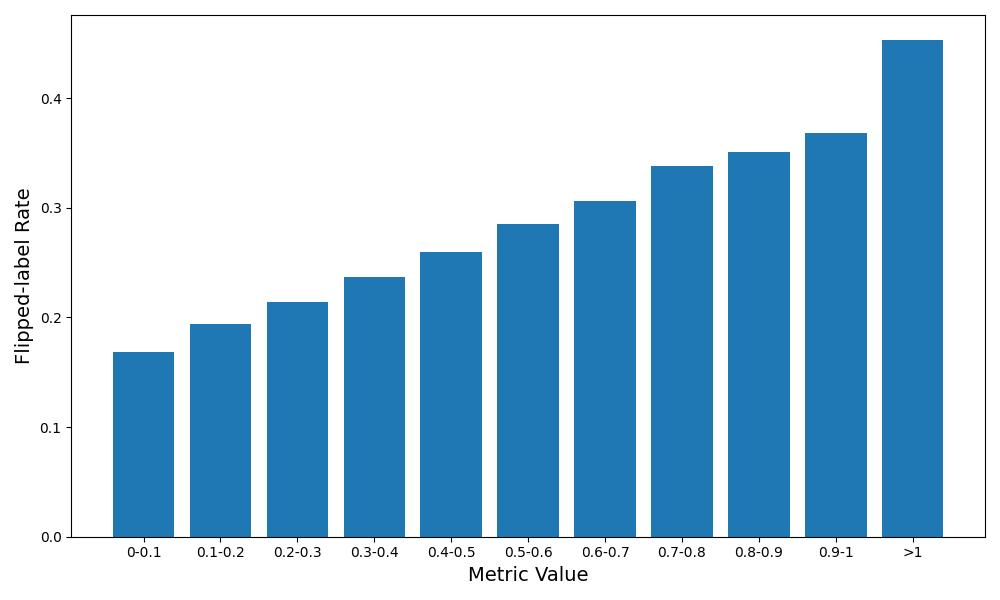}
\end{center}
\vspace{-0.15in}
 \caption{Here, we add 20\% label flip to Pick-a-Pic v2~\cite{kirstain2024pick} and  calculate the metric according to Eq.~\ref{eq:noise_aware_metric_dpo}. The x axis denotes the interval of the metric and the y axis denote the ratio of noisy samples. We can observes a significant increase of flipped sample ratio as the increase of the metric value.} 
 \label{fig:noise_rate}
\vspace{-0.2in}
\end{figure}

To combine the two terms together, we choose to directly use the product of them, which can be formulated as:
\begin{equation}\label{eq:noise_aware_metric_dpo}
u_\theta(x) = s_\theta(x)\cdot c_\theta(x)
\end{equation}

For our metric, the larger value $u_\theta(x)$ has, the higher likelihood the corresponding has to be a minority labeled one.
To understand the mechanism, we give a more detailed discussion. When the value is quite large, either $s_{\theta}(x)$ or $c_{\theta}(x)$ will be large which means the large value of our metric is related to detecting the pattern of large \textit{bias} or large \textit{variance}. For the sample with small  \textit{bias} and \textit{variance}, this value tends to be much smaller. And such kind of samples would be considered as majority sample during the training.

Previous methods~\cite{goel2022pars} mainly take the straightforward method that directly re-labels a part of the dataset and train a binary classifier based on these data. Then the two terms can be defined according to the output of the classifier. However, re-labeling the preference data by human annotator wastes a lot of manpower and material resources, thus being almost impossible for larger datasets. Moreover, such an annotation process would again introduce new biased labels. Besides, the binary classifier would require specific design for different input modalities such as images and texts, thus introducing extra complexity. Compared with that, only the finetuning of the target model is required without any extra models for our method.

To validate the efficacy of our metric, we conduct a pilot experiment. In detail, we manually add flipped label as minority samples to the preference data. Afterwards, we visualize the ratio the flipped samples with different values of the metric. The figure is shown in Figure.~\ref{fig:noise_rate}. By observation, we can find that with the increase of the metric, the ratio of flipped samples also increases accordingly. It can validate our metric as a first step. In the experiments we will show that the proposed metric is also effective when dealing with minority data in real scenarios.

\subsection{Adaptive-DPO}

In order to utilize the minority-instance-aware metric $u_{\theta}$ to enhance DPO, we mainly follow two intuitions: (1) making minority samples less important and (2) amplifying supervision from the majority samples. Given that larger value of $u_{\theta}$ is more likely to represent that the preference data is labeled with minority preference, we hence introduce both a weighting coefficient and an adaptive margin term to DPO objective according to the metric as follows,
\begin{equation}
\begin{aligned}
    \mathcal{L}_{Adaptive-DPO}(\pi_{\theta};\pi_{ref}) &= \\
    -\mathbbm{E}_{c,x^{w},x^{l}}
    \big\{W_{\theta}(x)& \big[\log\sigma \big( \beta  \ell_{\theta}(x)- \Gamma_{\theta}(x)\big)\big]\big\}\label{eq:adative-dpo}
\end{aligned}
\end{equation}
\begin{align}
W_{\theta}(x) &= sg(\frac{1}{1 + k_1 u_\theta(x)} ) \label{eqW}\\
    \Gamma_{\theta}(x) &= sg(k_2 u_\theta(x)^2 +c_2 ) \label{eqGa} 
\end{align}
$sg$ denotes stop gradient. The effect of this weighting term is intuitive. Large $u_{\theta}$ indicates a sample pair is likely to be labeled as minority, its corresponding $W_{\theta}$ will be relatively small, thus weakening the supervision from this sample. On the other hand, similar to SimPO~\cite{meng2024simpo}, the margin term can promote the generalization ability of the fine-tuned model. While SimPO relies on tuning the margin as a hyper-parameter, the margin $\Gamma_{\theta}$ introduced by us is adaptive according to the minority-instance-aware metric $u_{\theta}$. When $k_2 < 0$, smaller $u_{\theta}$ will be accompanied with larger $\Gamma_{\theta}$, \textit{i.e.} the objective can encourage model to produce more confident prediction with regard to majority samples. On the contrary, for those minority samples, the supervision induced by margin would be negligible, thus being fully controlled by the former weighting coefficient $W_{\theta}$. We provide further understanding of our loss in the perspective of the gradient with following form:
\begin{equation}\label{eq:grad-start}
\begin{aligned}
    & \nabla_{\theta}\mathcal{L}_\text{Adaptive-DPO}(\pi_{\theta}; \pi_{ref})  \\
    &= -\nabla_{\theta}\mathbb{E}_{c,x^{w},x^{l}} \left\{ W_{\theta}(x) \log\sigma \left( \beta \ell_{\theta}(x) - \Gamma_{\theta}(x) \right) \right\} \\
    &= -\mathbb{E}_{(x, y_w, y_l) \sim \mathcal{D}} \left[ \beta W_{\theta}(x) \sigma \left( -\beta \ell_{\theta}(x) + \Gamma_{\theta}(x) \right) \nabla_\theta \ell_\theta(x) \right]
\end{aligned}
\end{equation}
For the  pair that is more likely to contain minority preference, $u_\theta(x)$ tend to be  larger. 
As a consequence $W_{\theta}(x)$ and $\Gamma_{\theta}(x)$  tend to be smaller.
The first term will downweight this pair to alleviate the potential negative effect to training.
By viewing Eq.~\ref{eq:grad-start}, we find that if we find a suitable value of $k_2$ and $c_2$, the part $\sigma \left(-\beta * \ell_{\theta}(x)+ \Gamma_{\theta}(x) \right)$ will be suppressed accordingly which makes this pair less useful in that step. For clearance, we summarize the whole training process  in Alg.~\ref{alg:Reweight}.

\begin{algorithm}[h]
  \caption{Adaptive-DPO Training} \label{alg:Reweight}
  \small
  \begin{algorithmic}[1]
    \REQUIRE Pairwise preference dataset $ D = \{x^{(i)} = (x_{(i)}^w,x_{(i)}^l,c_{(i)})\}_{i=1}^{N}$
    \ENSURE Target model
    \FOR{batch in $D$}
      \FOR{$x^{(i)}$ in batch}
        \STATE Calculate $\ell_{\theta^{(m)}}(x)$ using (\ref{ltheta}) 

        \STATE Calculate $u_{\theta}$ by (\ref{eq:noise_aware_metric_dpo}) using $\{ \ell_{\theta^{(m)}}(x)\}_{m=1}^{M}$
        \STATE Calculate $W_{\theta}$ and $\Gamma_{\theta}$ by (\ref{eqW}) and (\ref{eqGa})
      \ENDFOR
      \STATE Optimize model with loss (\ref{eq:adative-dpo})
    \ENDFOR
  \end{algorithmic}
\end{algorithm}

\section{Experiments}
\subsection{Experiment Setup}

\noindent\textbf{Dataset.} In order to show the generalizability of our proposed method, we carry out experiments on the text-to-image generation task with both SD1.5 and SDXL models. We follow Diffusion-DPO to adopt Pick-a-Pic v2~\cite{kirstain2024pick} for training, which consists of 959k preference data. For evaluation, 500 unique test prompts and 500 unique validation prompts from Pick-a-Pic v2 are utilized. We also use extra test dataset HPDv2~\cite{wu2023human} which contains 400 test prompts.

\noindent\textbf{Evaluation protocol.} Our experiments consist of two main parts. First, we conduct experiments using the same flip-label setting as in the pilot study to verify that our metric is indeed effective for synthetic minority samples. SD1.5 is adopted for this experiment. Then we train models on the original Pick-a-Pic v2 with the proposed Adaptive-DPO and evaluate our method on other real-world benchmark to both: 1) support our claim that the existing preference datasets have the problem of minority data and 2) show that our method has great practical value in real-life alignment usage. For this part, both SD1.5, and SDXL are used for fine-tuning. We adopt PickScore~\cite{kirstain2024pick}, ImageReward~\cite{xu2024imagereward}, Aesthetic score~\cite{schuhmann2022laion} , and HPS~\cite{wu2023better} as evaluation metrics. 

\noindent\textbf{Implementation details.} All the experiments are launched by a total batch size of 128,  and re-scale parameter $\rho$ in \ref{eq:confidence_dpo} being 15 to ensure that the scale of $\ell_\theta (x)$ in Eq.~\ref{eq:noise_aware_metric_dpo} contains consistency in scale. DPO parameter $\beta$ in Eq.~\ref{eq:adative-dpo} is set to 1000 for SD1.5 and 2500 for SDXL. For hyper-parameters in Eq.~\ref{eq:noise_aware_metric_dpo} , $k_1$ set to 10 for SD1.5,and 1 for SDXL; $k_2$ set according to $\beta$ and $c_2$ according to the mean value of logits in each batch to be consistent with the scale of logits in loss for each model.

\begin{table}[htb]
\small
\parbox{\linewidth}{
\caption{Diffusion-DPO results with different label-flip rate on SD1.5. For all metrics, the larger value indicates the model is better. We copy results of DPO from Tab.~\ref{tab:pivot2} for better comparison and understanding.}
\centering
\resizebox{\linewidth}{!}{%
\begin{tabular}{cccccc}
      \toprule
      Method & Label-flip rate (\%) & IR ($\uparrow$) & PS ($\uparrow$) & Aes ($\uparrow$) & HPS ($\uparrow$)  \tabularnewline
      \midrule
        \multirow{3}{*}{DPO} & 10 & 0.05 & 20.91 & 5.27 & 26.32 \tabularnewline
        & 20 & 0.00 & 20.83 & 5.24 & 26.24 \tabularnewline
        & 30 & -0.05 & 20.72 & 5.21 & 26.14 \tabularnewline
        \midrule
        \multirow{3}{*}{Ours} & 10 & 0.39 & 21.40 & 5.43 & 26.77 \tabularnewline
        & 20 & 0.34 & 21.31 & 5.44 & 26.66 \tabularnewline
        & 30 & 0.31 & 21.26 & 5.41 & 26.50 \tabularnewline
      \bottomrule  
      \label{tab:synthetic}
    \end{tabular}
    }
\vspace{-0.15in}
}
\end{table}

\begin{table*}[t]
\parbox{\linewidth}{
\small
\caption{Quantitative results on SD1.5 and SDXL. The larger metric indicates the model is better. }
\vspace{-0.1in}
\centering
\resizebox{\linewidth}{!}{%
\begin{tabular}{cccccccccccccc}
\toprule
        &       &   \multicolumn{4}{c}{Pick-a-Pic Valid} & \multicolumn{4}{c}{Pick-a-Pic Test} & \multicolumn{4}{c}{HPDv2}\tabularnewline
   Backbone & Method &  IR ($\uparrow$)   &  PS ($\uparrow$)     &  Aes ($\uparrow$)  & HPS ($\uparrow$)  &  IR ($\uparrow$)   &  PS ($\uparrow$)     &  Aes ($\uparrow$)  & HPS ($\uparrow$)  &  IR ($\uparrow$)   &  PS ($\uparrow$)     &  Aes ($\uparrow$)  & HPS ($\uparrow$) \tabularnewline
    \midrule
    \multirow{5}{*}{SD1.5} & Pretrain       &  -0.15 &  20.58 &  5.16 & 26.02 &   -0.15& 20.66  &  5.32 &  26.12  & -0.12  & 20.90  &   5.24& 26.72 \tabularnewline
    &Diffusion-DPO                      &0.16 & 21.05 & 5.31 & 26.43   &  0.23 &\underline{21.23}   & \underline{5.53}  &  26.61  &  0.19 & 21.41  & 5.42  & 27.17   \tabularnewline
    &Robust-DPO                         & 0.20  &  \underline{21.13} & \textbf{5.51}  &  26.68  & 0.24 & 21.21   & 5.52  & 26.64  & 0.30  & \underline{21.58}  & \textbf{5.68} & 27.29 \tabularnewline
    &SFT$_{chosen}$                      &  \underline{0.31} &  20.88 &  5.40 & \textbf{26.79}  &  \underline{0.33}  & 20.98  & \underline{5.53}  & \underline{26.76} &   \underline{0.42} &  21.39 & 5.54  & \underline{27.53}  \tabularnewline
    &Ours                               & \textbf{0.43} & \textbf{21.35} & \underline{5.44} & \underline{26.76}   & \textbf{0.42}  &  \textbf{21.43} &  \textbf{5.57} &   \textbf{26.86} &\textbf{0.53}   &  \textbf{21.84} &  \underline{5.58} & \textbf{27.57} \tabularnewline
    \midrule
    \multirow{5}{*}{SDXL} &  Pretrain     &  0.51 & 22.10 & 5.86  &  26.80  &  0.57 &  22.16 &  6.01 &  26.79  &  0.76 &  22.84 &  6.14  &  27.49\tabularnewline
    &Diffusion-DPO                   &  0.87 & 22.62 & 5.89  &  27.32   & 0.86  &  22.66 &  6.02 &  27.27    &  0.99 & 23.27  &  6.13 &  27.96   \tabularnewline
    &Robust-DPO                         &  \underline{0.93} & \underline{22.66}  & \underline{6.01}  & \textbf{27.47}  &  \underline{0.90} & \underline{22.73}  & 6.01  & \underline{27.36}   &  \underline{1.09} &  \underline{23.34} &  6.13 & \underline{28.06}  \tabularnewline
    &SFT$_{chosen}$                     &  0.59 &  22.35 & \textbf{6.09}  & 26.80   &  0.61 & 22.40  & \textbf{6.08}  & 26.73    &  0.81 & 23.04  &  \textbf{6.18} &  27.34 \tabularnewline
    &Ours                               & \textbf{0.95} & \textbf{22.77} & 5.94 & \underline{27.46} & \textbf{0.95} & \textbf{22.82} & \underline{6.06} & \textbf{27.37}  & \textbf{1.10} & \textbf{23.40} & \underline{6.17} & \textbf{28.09}  \tabularnewline
\bottomrule
\end{tabular}
\label{tab:real-diffusion}
}
\vspace{-0.15in}
}
\end{table*}
 
\subsection{Experiments on Synthetic Minority Data}\label{sec:synthetic}

To validate the effectiveness of our method, we first conduct a simple experiment based on synthetic minority samples. Concretely, following the operation in Sec.~\ref{sec:pivot2}, given a specific proportion, part of the training samples are randomly chosen and their corresponding preference labels are flipped. Then the data is used to fine-tune the pretrained SD1.5 via our proposed Adaptive-DPO. In this way, we can preliminarily verify the effectiveness of the method. 

As shown in Tab.~\ref{tab:synthetic}, with Adaptive-DPO, the fine-tuned model enjoys significantly stronger performance than the one fine-tuned with original DPO at the same label-flipped level. Compared with experiment below in Tab.~\ref{tab:real-diffusion}, we will see that even compared with model fine-tuned with real data and no synthetic noise, our Adaptive-DPO working on different label-flipped level still turns out to be better than Diffusion-DPO. Moreover, as the label-flipped rate gets larger, our method can to some extent combat against the performance degradation resulted from minority samples, demonstrated by the smaller performance decrement. This reflects the efficacy of our method against such synthetic minority data.

The improvement can be attributed to the learning process of the model. In general, during training, the model tends to first learn the information in the majority label, so as to continuously improve the prediction ability of its implicit reward model as in Eq.~\ref{eq:main_eq}. In this way, $c_{\theta}$ as in Eq.~\ref{eq:confidence_dpo} will be more credible and more accurate during the training process as the model being improved, thus the whole metric can get more credible. Consequently, the metric can function better and help the model eliminate the negative effect of minority samples.

\begin{figure*}[t]
    \centering
    \includegraphics[width=0.95\linewidth]{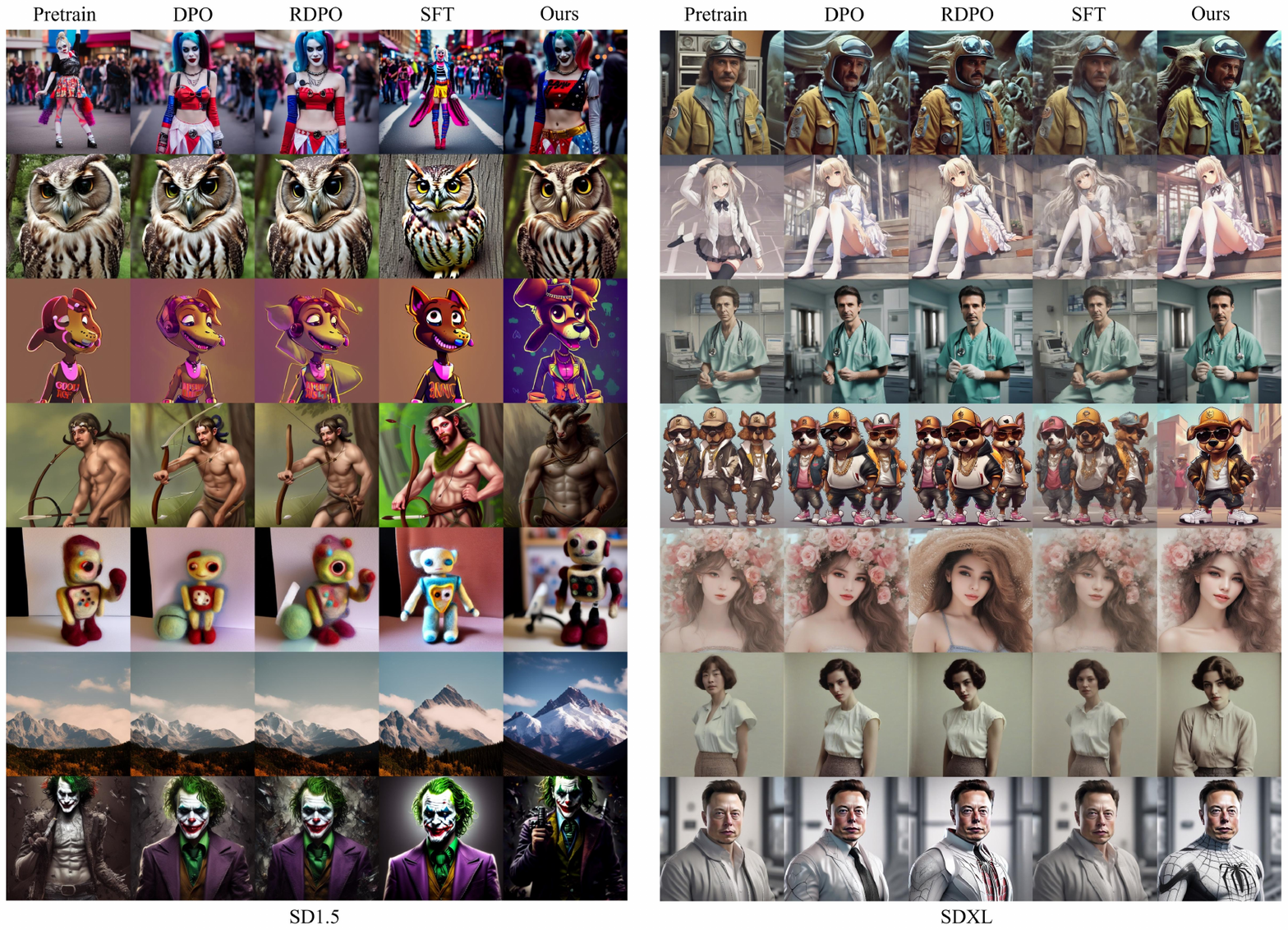}
    \vspace{-0.15in}
    \caption{Qualitative results with SD1.5 and SDXL as backbone. Please refer to supplementary for the corresponding prompts.}
    \vspace{-0.1in}
    \label{fig:main}
\end{figure*}

\subsection{Experiments on Real Data}

As mentioned in Sec.~\ref{sec:pivot2}, the above experiments on synthetic flipped labels cannot fully present the real value of our method, since the distribution of minority data created during the annotation process can be significantly different from that of the synthetic flipped labels. 
To this end, we directly fine-tune the pretrained models on Pick-a-Pic v2 datasets with our proposed Adaptive-DPO and other comparison methods including Diffusion-DPO, Robust-DPO, and SFT with chosen samples in data pairs. The results are provided in Tab.~\ref{tab:real-diffusion}. 

One can find that even there is no synthetic flipped label, applying our method is still better than Difsuion-DPO in all these different settings. Specifically, for SD1.5, our method can outperform DPO in terms of ImageReward by 0.21, with consistent results for other settings. Additionally, the Robust-DPO method, while achieving state-of-the-art performance in the noisy domain and possessing substantial theoretical support, demonstrates suboptimal training results on real-world datasets. This indicates that the issue we need to address which is the minority preference problem within preference data, cannot be resolved through simply using the existing Learning with Noise Labels (LNL) methods. Our method, different from LNL-based Robust-DPO, is not only effective for the synthetic noise, but also for the real-case noise, thus indicating the generalization ability of our method.

For clearer comparison, we in Fig.~\ref{fig:main} visualize some image results generated by different methods and backbones. While SFT generally enjoys good quantitative performance, the generated images are in low quality. Compared with the images generated by Diffusion-DPO, our results generally enjoy better quality, which are consistent with the quantitative results. Specifically, for SD1.5, model fine-tuned with our method can generate more details both for human bodies and backgrounds. As for SDXL, we find that while the pretrained model and DPO finetuned one tend to generate mis-located limbs and deformed hands, which lowers the image quality, our method can help alleviate this problem, resulting in more delicate human portraits. 

\subsection{Ablation study}\label{sec:ablation}
To further validate the effectiveness of our method, we conduct a series of ablation study regarding the design of our objective and several hyperparameters. If not specified, we present results on Pick-a-Pic validation set using SD1.5 as backbone.

\begin{table*}[t]
\parbox{\linewidth}{
\small
\caption{Comparison of Adaptive-DPO and voting strategy. For all metrics, the larger value indicates the model is better. }
\vspace{-0.1in}
\centering
\resizebox{\linewidth}{!}{%
\begin{tabular}{cccccccccccccc}
\toprule
        &       &   \multicolumn{4}{c}{Pick-a-Pic Valid} & \multicolumn{4}{c}{Pick-a-Pic Test} & \multicolumn{4}{c}{HPDv2}\tabularnewline
   Backbone & Method &  IR ($\uparrow$)   &  PS ($\uparrow$)     &  Aes ($\uparrow$)  & HPS ($\uparrow$)  &  IR ($\uparrow$)   &  PS ($\uparrow$)     &  Aes ($\uparrow$)  & HPS ($\uparrow$)  &  IR ($\uparrow$)   &  PS ($\uparrow$)     &  Aes ($\uparrow$)  & HPS ($\uparrow$) \tabularnewline
    \midrule
    \multirow{3}{*}{SD1.5} &Diffusion-DPO                      &0.16 & 21.05 & 5.31 & 26.43   &  0.23 & 21.22   & 5.53  &  26.61  &  0.19 & 21.41  & 5.42  & 27.17   \tabularnewline
    & Voting+Diffusion-DPO       &  0.32 &  21.28 &  5.40 & 26.62 &  0.35 & 21.38  &  \textbf{5.58} &  26.78  & 0.46  & 21.80  &   5.53 & 27.45 \tabularnewline
    &Ours                               & \textbf{0.43} & \textbf{21.35} & \textbf{5.44} & \textbf{26.76}   & \textbf{0.42}  &  \textbf{21.43} &  5.57 &   \textbf{26.86} &\textbf{0.53}   &  \textbf{21.84} &  \textbf{5.58} & \textbf{27.57} \tabularnewline
\bottomrule
\end{tabular}
\label{tab:ablation-voting2}
}
\vspace{-0.2in}
}
\end{table*}

\noindent\textbf{Is there any straightforward alternatives for solving the minority samples problem?} One would ask if it is necessary to design such a method to deal with the problem, and if there is any other simpler solution, such as using training data with higher quality. To answer this question, we in this part adopt a majority voting strategy to re-filter the training data. Concretely, we use PickScore~\cite{kirstain2024pick}, ImageReward~\cite{xu2024imagereward}, Aesthetic score~\cite{schuhmann2022laion}, and HPS~\cite{wu2023better} to re-annotate the whole Pick-a-Pic v2. In this way, along with the original human annotate results, we get 5 labels for each data pair. Then these data are re-annotated according the majority of the 5 labels and utilized to finetune the pretrained SD1.5 using Diffusion-DPO. The results in Tab.~\ref{tab:ablation-voting2} show that majority voting to some extend can enhance the confidence of the datasets, but its performance is not as good as our approach. Furthermore, majority voting with learned scores needs to re-annotate the datasets for several times to select the majority label of every sample, which is less efficient than our Adaptive-DPO considering labor and computation cost as well as convenience for training, not to mention re-annotating with human sources.

\begin{table}[htb]
\small
\parbox{\linewidth}{
\caption{Ablation study results among different variants regarding objective design. For all metrics, the larger value indicates the model is better. For the specific formulas of different variants please refer to supplementary material. }
\vspace{-0.1in}
\centering
\resizebox{\linewidth}{!}{%
\begin{tabular}{ccccc}
      \toprule
      Variants & IR ($\uparrow$) & PS ($\uparrow$) & Aes ($\uparrow$) & HPS ($\uparrow$)  \tabularnewline
      \midrule
        w/o margin & 0.41  & 21.35 & 5.41 & 26.71 \tabularnewline
        + linear margin & 0.42 & 21.35   & 5.43 & 26.75 \tabularnewline
        + quadratic margin (Ours) & \textbf{0.43} & \textbf{21.35} & \textbf{5.44}  & \textbf{26.76} \tabularnewline
        \midrule
        square re-weight &  0.37  &  21.34 & 5.61  &  26.90 \tabularnewline
        sqrt re-weight &  0.21 &  20.86 &  5.33 &   26.45\tabularnewline
        sigmoid re-weight &   0.07   &  20.75 &  5.22 & 26.28  
        \tabularnewline
        linear re-weight (Ours) & \textbf{0.43} & \textbf{21.35} & \textbf{5.44}  & \textbf{26.76} \tabularnewline
      \bottomrule  
      \label{tab:ablation-objective}
    \end{tabular}
\vspace{-0.2in}
    }
}
\end{table}

\begin{figure}[t]
    \centering
    \includegraphics[width=0.85\linewidth]{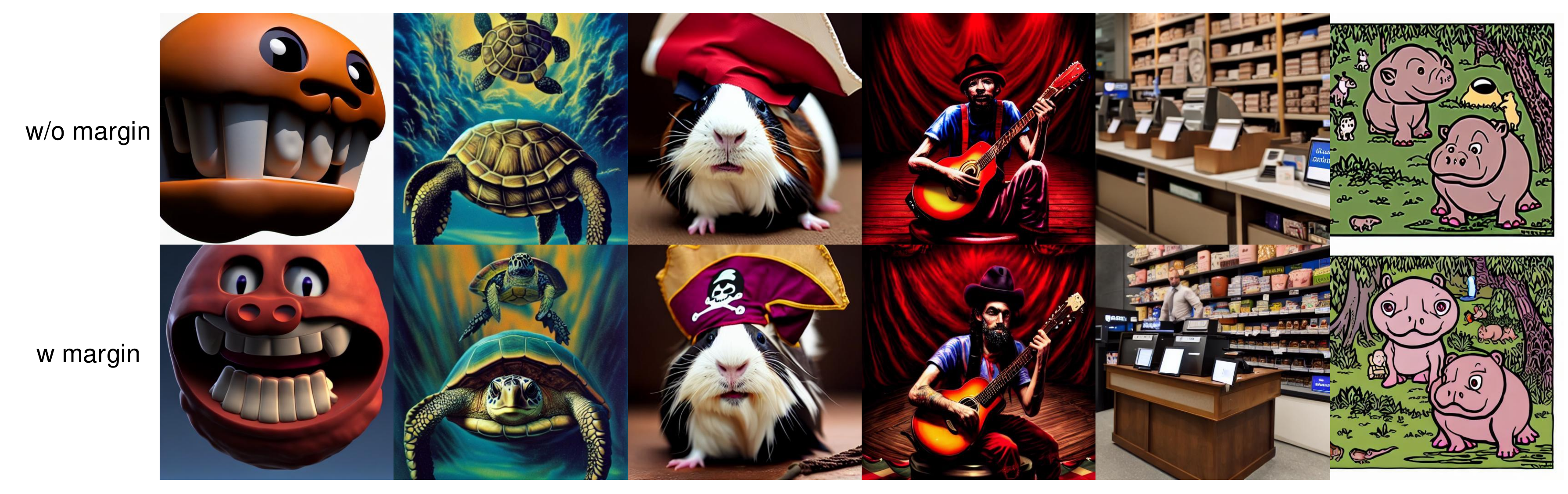}
    \vspace{-0.1in}
    \caption{Qualitative comparison between models finetuned with and without the proposed margin term.}
    \label{fig:ablation-margin}
    \vspace{-0.15in}
\end{figure}

\noindent\textbf{Effectiveness of the Adaptive-DPO objective.} We try to analyze the design of our proposed objective, for which model variants without the proposed adaptive margin, with linear and quadratic margin, as well as with square, sqrt, sigmoid and linear re-weight are compared. The results are shown in Tab.~\ref{tab:ablation-objective}. Other re-weighting methods, such as square, sqrt, and sigmoid, result in lower performance, highlighting the importance of our design. Meanwhile, the quadratic margin, our proposed method, achieves the highest scores in all metrics, demonstrating its effectiveness in enhancing the model's capability, which can also be validated in the qualitative results in Fig.~\ref{fig:ablation-margin}. Since the proposed margin term works by enhancing the supervision from the majority samples, its impact is reasonably weaker than the reweighting term which can eliminating the negative effect brought by minority samples.

\noindent\textbf{Role of different hyperparameters.} We conduct the sensitivity studies regarding hyperparameters in this paragraph. Tab.~\ref{tab:ablation-hp} shows that our method is relatively robust to the change of hyper-parameters. Specifically, with $k_1$ increasing from 4 to 10, the ImageReward increases by 0.05 while PickScore gets lower. When $k_1$ is larger than 10, the performance generally saturates, which may be attributed to the overfitting problem. Apart from the above parameter, $\rho$ in Eq.~\ref{eq:std_dpo} also has an impact on the results. Larger $\rho$ can make the difference in the given implicit label between the minority and majority samples greater, resulting in more obvious improvement in training.

\begin{table}[t]
\small
\parbox{\linewidth}{
\caption{Ablation study results among different variants regarding hyperparameter value. For all metrics, the larger value indicates the model is better.}
\vspace{-0.1in}
\centering
\resizebox{\linewidth}{!}{%
\begin{tabular}{ccccccc}
      \toprule
      Hyperparameter & Value & IR  & PS  & Aes  & HPS  \tabularnewline
      \midrule
       \multirow{3}{*}{$k_1$} & 4 & 0.38  & 21.36 & 5.44 & 26.75 \tabularnewline
       & 10 & 0.43 & 21.35 & 5.44 & 26.76  \tabularnewline
       & 12 & 0.42 & 21.33 & 5.41  & 26.73 \tabularnewline
    \midrule
     \multirow{2}{*}{$\rho$} & 10 &  0.39 &  21.39  &  5.47 & 26.72 \tabularnewline
       & 15 & 0.43 & 21.35 & 5.44 & 26.76  \tabularnewline
      \bottomrule       
      \label{tab:ablation-hp}
    \end{tabular}
    }
}
\vspace{-0.3in}
\end{table}

\noindent\textbf{Application of our method to other baseline.} To show the generalization ability of our method, we select a strong baseline IPO~\cite{azar2024general} and extend our method to its formulation, named as Adaptive-IPO. The results are shown in Tab.~\ref{tab:ablation-dpo}. It is obvious that with such a strong baseline, adopting our method can still lead to significant improvement. This further indicates that our method is a generalizable remedy for DPO and its followers to solve the problem of minority data in the training set.

\begin{table}[t]
\small
\parbox{\linewidth}{
\caption{Ablation study results using IPO as baseline. For all metrics, the larger value indicates the model is better.}

\vspace{-0.1in}
\centering
\resizebox{\linewidth}{!}{%
\begin{tabular}{ccccc}

      \toprule
      Method & IR ($\uparrow$) & PS ($\uparrow$) & Aes ($\uparrow$) & HPS ($\uparrow$)  \tabularnewline
      \midrule
       IPO &  0.16 &  21.09 & 5.31  & 26.48 \tabularnewline
       IPO+Ours & \textbf{0.33}  & \textbf{21.17}  &\textbf{5.36} & \textbf{26.58} \tabularnewline
      \bottomrule  
      \label{tab:ablation-dpo}
    \end{tabular}
    }
\vspace{-0.2in}
}
\end{table}

\section{Related Work} 
Due to space limitations, we refer the reader to the supplementary materials for full related works.

\noindent\textbf{Aligning large-scale models.} Recent advancements in aligning large-scale models with human preferences have introduced efficient alternatives to RLHF. DPO~\cite{rafailov2024direct} bypasses the need for reward models, while IPO~\cite{azar2024general}, KTO~\cite{ethayarajh2024kto}, and RPO~\cite{yin2024relative} extend preference optimization to various data settings. Robust-DPO~\cite{chowdhury2024provably} addresses noisy data but struggles with real-world applications. Our method effectively handles real-world preference data. 

\noindent\textbf{Aligning diffusion models.} The alignment of diffusion models has seen significant progress, with DDPO~\cite{black2023training} modeling image denoising as a Markov decision process. DPOK~\cite{fan2023dpok}, D3PO~\cite{yang2024using}, and Diffusion-DPO~\cite{wallace2024diffusiondpo} further optimize these approaches. Diffusion-KTO~\cite{li2025aligning} and Diffusion-RPO~\cite{gu2024diffusion} extend these methods to unpaired data. However, addressing subjectivity and uncertainty in preference annotations remains an open challenge, which our work tackles. 

\section{Conclusion}

This paper discusses the minority preference problem in preference data for the first time, and gives an experimental test to verify its existence and its negative influence on the model. We hence propose Adaptive-DPO as a solution to this problem. Adaptive-DPO not only demonstrates strong resistance to artificially introduced minority samples, achieving reasonably normal fine-tuning performance even at flipped-label rate as high as 30\%, but also shows significant improvement when applied to real-world data, effectively addressing the issue of minority preference in naturally annotated datasets. Furthermore, Adaptive-DPO can be successfully applied across various models and derivative methods, such as IPO, showing versatility in both SD1.5 and SDXL. The improvements observed in these areas highlight its potential to enhance performance in a wide range of applications. 

{
    \small
    \bibliographystyle{ieeenat_fullname}
    \bibliography{main}
}

\clearpage  

\begin{center}
    {\LARGE \textbf{Supplementary Materials}}  
\end{center}
\bigskip  

\appendix

\section{Discussion}

\paragraph{Relationship with previous works.} One would note that there are several papers working on noisy data during finetuning LLMs with DPO, such as Robust-DPO~\cite{chowdhury2024provably} and Gao et. al.~\cite{gao2024impact}. We argue that their scopes are different from ours. For text data, the preference criteria is relatively simple, e.g. length or accuracy. Therefore, the noisy samples studied in the previous works mainly belong to the erroneous annotations. Consequently, \cite{gao2024impact} proposed detecting the noisy sample with only confidence-based metric. Different with that, for image data we further consider the minority samples caused by subjectivity divergence, which however cannot be reasonably detected by the previous confidence-based metric. To this end we propose the Adaptive-DPO with the novel minority-aware metric. Moreover, our method is self-adaptive to different proportion of minority samples in the training data, which is different from Robust-DPO requiring noisy levels.  

\paragraph{Future works.} Based on our proposed Adaptive-DPO, future work can focus on exploring more comprehensive analyses about the minority data in the preference data. Investigating the underlying principles of this approach could offer deeper insights into its mechanics, which may lead to more targeted enhancements and wider applicability across different domains. Understanding the theoretical foundations will also contribute to fine-tuning the parameter space for even better performance under different conditions, further reinforcing the method’s adaptability and robustness.

\section{Related Work}
\subsection{Aligning large scale models}
Due to the storage and computational limitation in RLHF, several alternative approaches have been proposed to effectively learn information from human preferences. First, the emergence of DPO \cite{rafailov2024direct}  allows for RLHF to bypass the need for training a reward model, enabling the model to directly learn from preference data, which makes the training process more convenient and efficient. Next, IPO \cite{azar2024general} introduced a new optimization approach that utilizes squared losses. Subsequently, KTO \cite{ethayarajh2024kto} expanded the application of preference optimization to unpaired data. Building on this foundation, RPO \cite{yin2024relative} made improvements that can be applied to both paired and unpaired data. In response to the presence of noisy data, Robust-DPO \cite{chowdhury2024provably} was proposed. However, we found that while Robust-DPO performs well on synthetic noisy data in the image generation domain, it does not yield improvements when applied to real-world data. Our method is equally effective for real-world data. 
 
\subsection{Aligning Diffusion Models}
Following the rapid development of large language models, the alignment of diffusion models with human preferences has also advanced significantly. DDPO \cite{black2023training} was the first to model the image denoising process as a multi-step Markov decision process, effectively applying reinforcement learning to the alignment of diffusion models. DPOK \cite{fan2023dpok} introduced KL loss and value function learning, further optimizing the methods of DDPO. D3PO \cite{yang2024using} applied DPO to diffusion models, making the alignment with human preferences more convenient. Meanwhile, Diffusion-DPO \cite{wallace2024diffusiondpo} further derived the loss for D3PO, directly optimizing the model's predicted noise within the loss, establishing it as the state-of-the-art method for applying DPO in diffusion models. After that, Diffusion-KTO \cite{li2025aligning} applied KTO to diffusion model, and Diffusion-RPO \cite{gu2024diffusion} applied RPO to diffusion model, making better use of the unpaired preference data in field of image generation. 
Also, Curriculum-DPO \cite{croitoru2024curriculum} aims to enhance the effectiveness of DPO by gradually introducing pairs that differ by increasingly subtle (fine-level) details during the training process. However, there has yet to be a proposed solution addressing the subjectivity and inherent uncertainty in the preference data annotation process, which is the issue this paper seeks to address.

\subsection{Relative Attribute}
Relative attribute \cite{parikh2011relative} is a concept that has been mentioned in the field of image classification. Distinguishing from binary attribute, it refers to image features that are subjective or difficult to assess, necessitating a richer representation beyond binary numerical values. In this era of large models, we find the learning of human preferences remains strongly correlated with the earlier proposed learning of relative attributes. Initially, relative attributes \cite{parikh2011relative} focused on class-level attribute comparisons, also there are some studies like \cite{fu2015robust} addressing outliers at the instance level. 
This paper aims to tackle the issue of minority preferences that inherently exist in crowd-sourced data. We focus on instance-level comparisons and propose a novel method for DPO regarding preference as a kind of relative attribute.

\subsection{Learning from noisy paired crowd-sourced data }

Training a more robust model using dataset with noisy labels is the target of learning with noisy labels. Methods employed are mostly in the field of image classification, including robust algorithm and noisy label detection.  Robust algorithm designs specific modules to ensure that the network can be well trained from the noise data set which includes the construction of robust networks such as~\cite{xiao2015learning, chen2015webly}, robust loss functions like  \cite{ghosh2017robust}.To give a solution to noisy data in DPO, Robust-DPO~\cite{chowdhury2024provably} uses a robust loss function to improve the model's resistance to the noisy label;  \cite{wang2024secrets} provided an insight of why noisy label influence reward model, and give their approaches to solve it. However, the distribution of paired crowd-sourced data is significantly different from image classification tasks, "correct" or "wrong" is hard to define sometimes as stated in Sec.1; the difference between preference annotations can be caused by either pure mistakes or annotators' subjectivity understanding. The former shows us noisy labels is indeed encompassed within the realm of minority preferences, while the latter factor makes it more complex than LNL.  

In this paper, we focus on real-world preference data and proposes an interpretable and effective method to clarify which label is more consistent with general public, rectifying the negative influence of minority preferences.

\section{Additional Results of Different Flipped-label Rate}
The experiments in the following Tab.\ref{tab:supp-flip} complement the synthetic minority experiment in Sec.5.2 of main paper, adding evaluation on HPDv2 and Pickapic v2 test dataset. We also train SD1.5 using Robust-DPO on different flipped-label level, which can be used as a further comparison. As can be seen from the comparison, our approach is much better than Robust DPO at different flip rates.

\begin{table*}[t]
\parbox{\linewidth}{
\small
\caption{Additional quantitative results of Diffusion-DPO, Robust-DPO and our method on different  flipped-label level with SD1.5. The larger metric indicates the model is better. }
\vspace{-0.1in}
\centering
\resizebox{\linewidth}{!}{%
\begin{tabular}{cccccccccccccc}
\toprule
        &       &   \multicolumn{4}{c}{Pick-a-Pic Valid} & \multicolumn{4}{c}{Pick-a-Pic Test} & \multicolumn{4}{c}{HPDv2}\tabularnewline
   Methods & Flip Rate(\%) &  IR    &  PS    &  Aes  & HPS   &  IR    &  PS    &  Aes & HPS   &  IR    &  PS      &  Aes  & HPS  \tabularnewline
    \midrule
    
    \multirow{3}{*}{Diffusion-DPO} & 10       &0.05 & 20.91 & 5.27 & 26.32  &0.09 & 20.99 & 5.44 & 26.40  & 0.12 & 21.29 & 5.38 & 27.05  \tabularnewline
    &20                     & 0.00 & 20.83 & 5.24 & 26.24  & 0.05 & 20.92 & 5.41 & 26.38   & 0.04 & 21.20 & 5.34 & 26.99 \tabularnewline
    &30                     &   -0.05 & 20.72 & 5.21 & 26.14   & -0.05 & 20.81 & 5.38 & 26.26   & -0.03 & 21.08 & 5.31 & 26.87\tabularnewline
     \midrule
    \multirow{3}{*}{Robust-DPO} & 10       & 0.11  & 21.01  & 5.45  & 26.56  & 0.10  & 21.02  & 5.45  & 26.45   & 0.17  & 21.37  & 5.59  & 27.14 \tabularnewline
    &20                     & 0.03  & 20.88  & 5.41  & 26.44  & 0.06  & 20.94  & 5.43  & 26.40   & 0.09  & 21.25  & 5.56  & 27.05 \tabularnewline
    &30                     &  -0.02 & 20.77  & 5.27  & 26.33  & -0.01  & 20.83  & 5.38  & 26.27   & -0.03  & 21.10  & 5.50  & 26.89 \tabularnewline
    \midrule
    \multirow{3}{*}{Ours} & 10       & 0.39 & 21.40 & 5.43 & 26.77  &0.39 & 21.44 & 5.59 & 26.85  &  0.49 & 21.84 & 5.59 & 27.51 \tabularnewline
    &20                     & 0.34 & 21.31 & 5.44 & 26.66& 0.36 & 21.34 & 5.59 & 26.77   & 0.47 & 21.79 & 5.60 & 27.43 \tabularnewline
    &30                     & 0.31 & 21.26 & 5.41 & 26.50  &0.26 & 21.16 & 5.58 & 26.64  &  0.31 & 21.60 & 5.54 & 27.22\tabularnewline
\bottomrule
\end{tabular}
\label{tab:supp-flip}
}
\vspace{-0.15in}
}
\end{table*}

\section{Formualtion of different variants of re-weight and margin}
We give below the formulas for the various vriants in Sec.5.4. Linear re-weight is as Eq.~\ref{eq:linear}; quardratic re-weight is as Eq.~\ref{eq:quardratic}; sqrt re-weight is as Eq.~\ref{eq:sqrt}; sigmoid re-weight is as Eq.~\ref{eq:sigmoid}. Linear margin is as Eq.~\ref{eq:linear-margin}, quardratic margin is as Eq.~\ref{eq:quardratic-margin}.

\begin{align}
W_{\theta}(x) &= \frac{1}{1 + k_1 u_\theta(x)}  \label{eq:linear}\\
W_{\theta}(x) &= \frac{1}{1 + k_1 u_\theta(x)^2} \label{eq:quardratic}\\
W_{\theta}(x) &= \frac{1}{1 + k_1 \sqrt{ u_\theta(x)}}  \label{eq:sqrt}\\
W_{\theta}(x) &= \frac{1}{1 + e^{k_1 u_\theta(x)}} \label{eq:sigmoid}\\
\Gamma_{\theta}(x) &= k_2 u_\theta(x) +c_2 \label{eq:linear-margin}\\
\Gamma_{\theta}(x) &= k_2 u_\theta(x)^2 +c_2 \label{eq:quardratic-margin}
\end{align}

\section{Details for Adaptive-IPO}

Based on the $\ell_{\theta}(x)$ in Eq.10, Adaptive-IPO loss can be written as:
\begin{equation}\label{eq:adaptive-ipo-loss}
\begin{aligned}
     &\mathcal{L}_{Adaptive-IPO}(\pi_{\theta};\pi_{ref}) = \mathbbm{E}_{c,x^{w},x^{l}} \\
    &\big[W_{\theta}(x) *\big( \ell_{\theta}(x)- \Gamma_{\theta}(x)-\frac{1}{2\beta}\big)^2 \big]
\end{aligned}
\end{equation}

\begin{figure}[htb]
    \centering
    \includegraphics[width=1\linewidth]{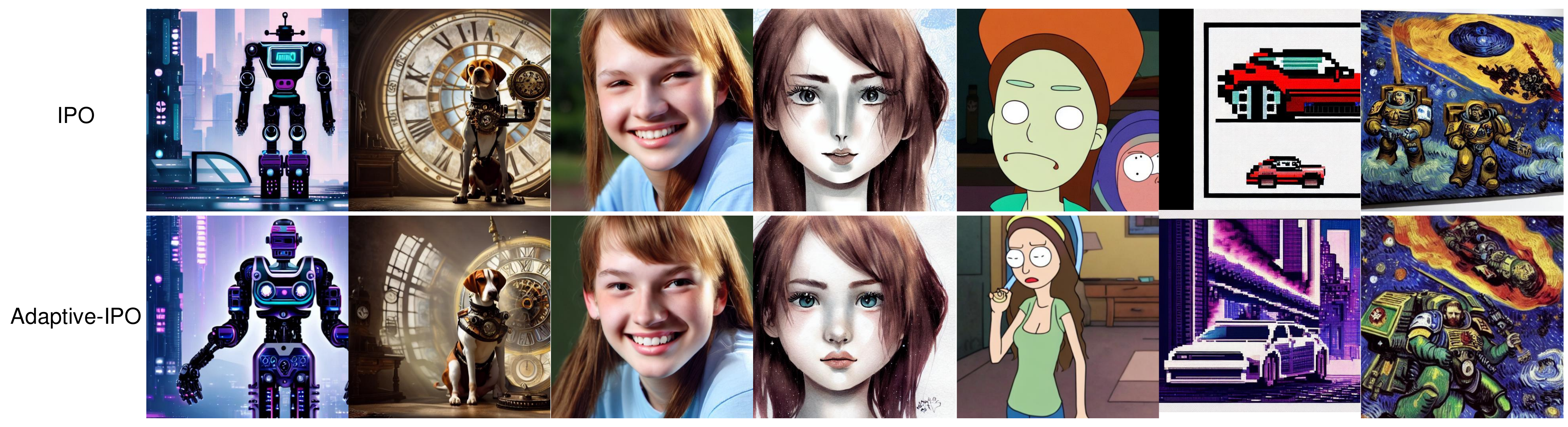}
    \caption{Qualitative comparison between using IPO and using Adaptive-IPO with SD1.5 as backbone.}
    \label{fig:supp-ipo}
\end{figure}

The qualitative results of using IPO as baseline are shown in Fig.~\ref{fig:supp-ipo}, which are consistent with those of using DPO as baseline.

\section{Prompts used for qualitative results}

\textbf{SD1.5}:
\begin{itemize}
    \item \textit{A fashion photograph of a Harley Quinn standing in the middle of a busy street, surrounded by a crowd of paparazzi, confident and poised, fashionable clothing, vibrant color, sharp lines and high contrast, 12k resolution, Canon EOS R5, natural lighting, 50mm lens.} 
\item\textit{a goofy owl.} 
\item\textit{a cute cartoon anthropomorphic african american insta baddie dog fursona wearing hip hop fashion and heels, trending on Artstation, gangster, vector drawing style, character design, style hybrid mix of patrick brown and kasey golden, dribbble 8k, airbrush concept art, full body, furry art.} 
\item\textit{a digital painting of a satyr archer.} 
\item\textit{a needle-felted robot.} 
\item\textit{a photograph of a mountain.} 
\item\textit{The joker holding pistols, vray, fantasy art, art by Russ Mills, blending, smooth, serious, detailed expressions, artstyle, detailed eyes, HDR, UHD, 64k, RTX, sharp, sharp focus, highly detailed, intricate detail, professional, artistic flow, ultra detailed, high resolution illustration
The gate to the eternal kingdom of angels, fantasy, digital painting, HD, detailed.} 
\end{itemize}

\noindent\textbf{SDXL}: 
\begin{itemize}
\item\textit{80's retrofuturism space-age, man as zoo keeper care about alien animal, very interesting movie set, beautiful clothes, insane details, ultra-detailed, extremely expressive body, photo portfolio reference, retrospective cinema, KODAK VISION3 500T, interesting color palette, cinematic lighting,  DTM, Ultra HD, HDR, 8K.} 
    \item \textit{a anime girl wearing white thighhighs.}  
\item\textit{a doctor wearing scrubs, holding a needle, staring at the camera.} 
\item\textit{a cute cartoon anthropomorphic african american insta baddie dog fursona wearing hip hop fashion and heels, trending on Artstation, gangster, vector drawing style, character design, style hybrid mix of patrick brown and kasey golden, dribbble 8k, airbrush concept art, full body, furry art.} 
\item\textit{a beautiful woman.} 
\item\textit{a photo of a woman.} 
\item\textit{analog style, face elon musk as like spiderman, 1080p, 16k Resolution, High Quality Rendering, RTX Quality, Realistic, Life Like. white background.} 
\end{itemize}

\section{Computation Cost of Adaptive-DPO}
Diffusion-DPO trained 12000 iterations on 8 V100 with a total batch size of 128 takes about 25 hours. The time taken by Adaptive-DPO is about 32 hours; For Majority Voting+DPO training, it takes about 10 hours to re-label the data with PickScore~\cite{kirstain2024pick}, ImageReward~\cite{xu2024imagereward}, Aesthetic score~\cite{schuhmann2022laion} , and HPS~\cite{wu2023better} on 8 V100, so the total time of data preprocessing and training is about 35 hours. In general, our method can achieve significnatly better results than DPO with majority voting with less time required.

\section{More Qualitative Results for Synthetic Data}
\begin{figure}[htb]
    \centering
    \includegraphics[width=\linewidth]{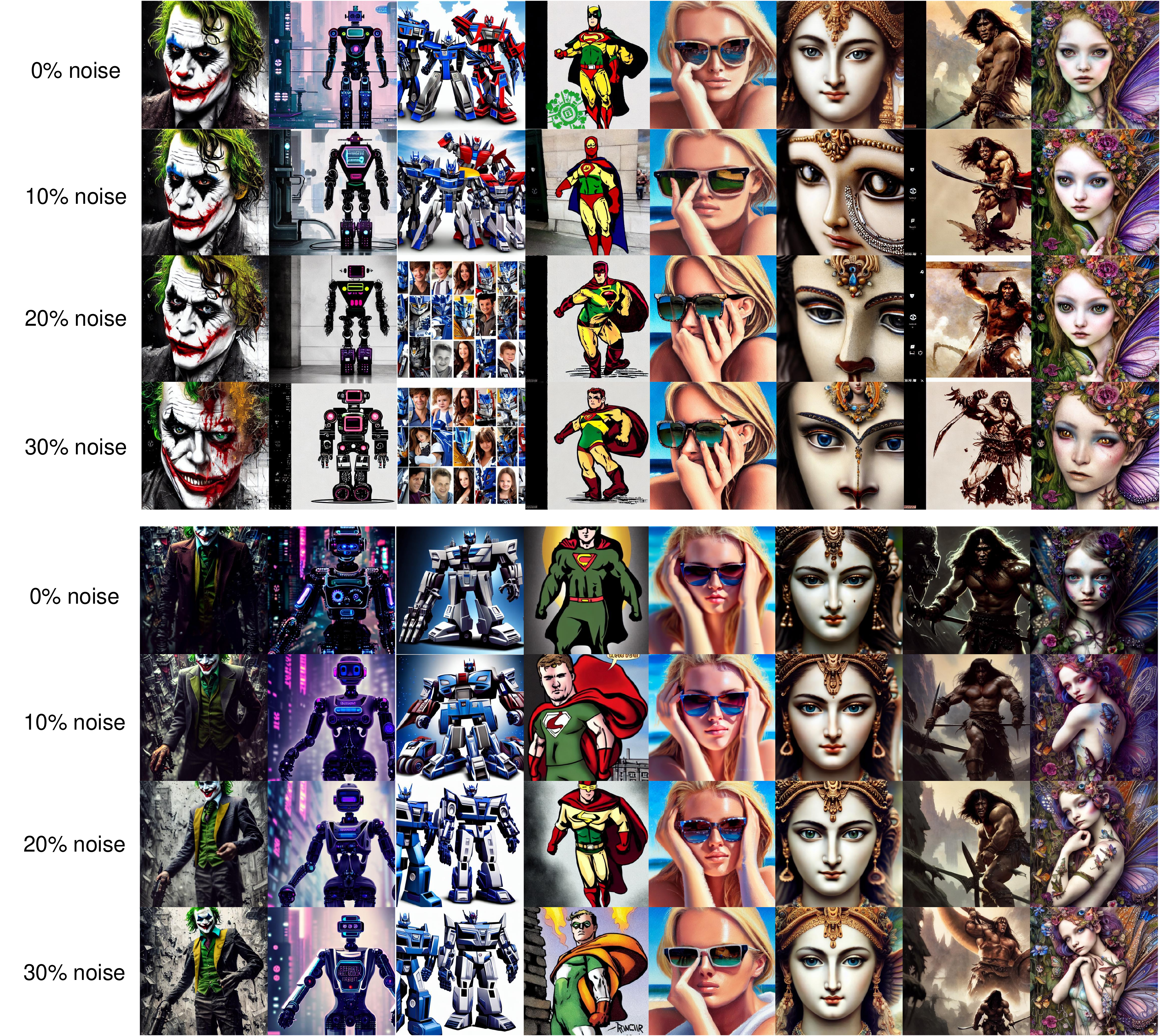}
    \caption{More qualitative comparison for Synthetic Data. The upper image grid denotes results generated by DPO under different noise rate, and the bottom one denotes ours. Note that the noise rate only represent the synthetic noise, but not the original noisy data contained in the training set.}
    \label{fig:supp-synthetic}
\end{figure}

\begin{figure*}[htb]
    \centering
    \includegraphics[width=1\linewidth]{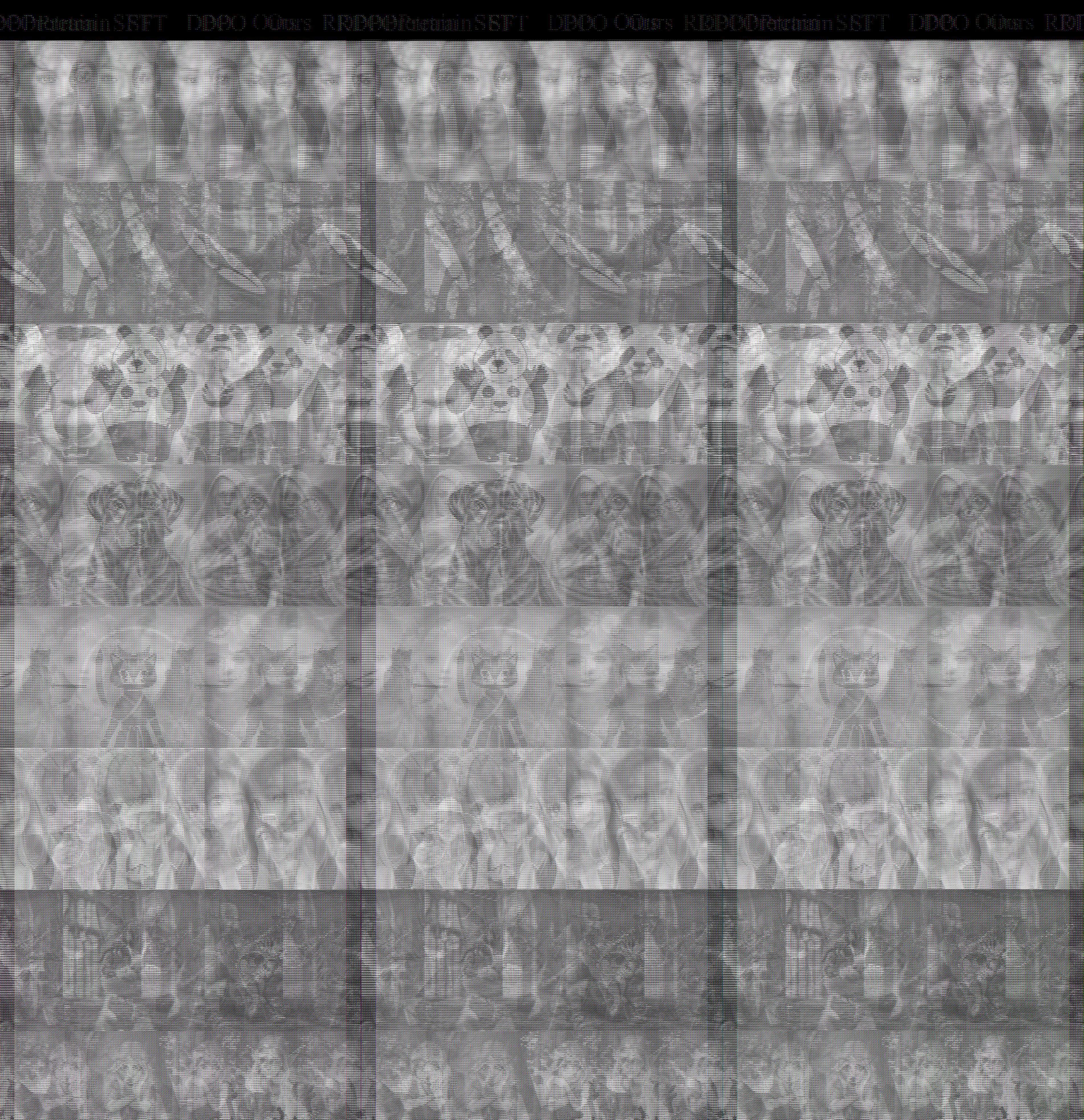}
    \caption{More qualitative comparison with SD1.5 and SDXL as backbone.}
    \label{fig:supp-results}
\end{figure*}
In Fig.~\ref{fig:supp-synthetic} we show some qualitative results for synthetic minority experiment. One can find that while the original DPO is vulnerable to the synthetic noise, our method is more robust against these data.

\section{More Qualitative Results for Real Data}

\begin{figure*}[htb]
    \centering
    \includegraphics[width=\linewidth]{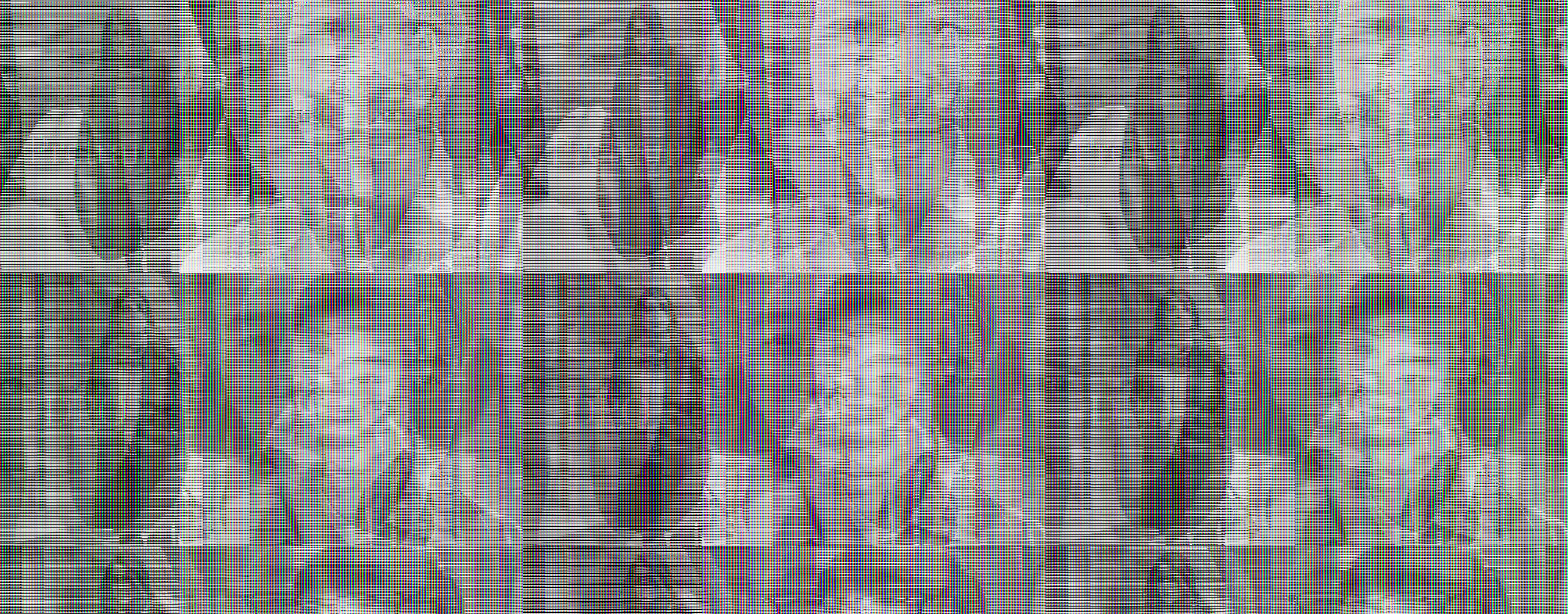}
    \caption{Additional results with SD1.5 as backbone training with portrait images.}
    \label{fig:supp-ablation}
\end{figure*}

In Fig.~\ref{fig:supp-results} we visualize more results including pretrained SD1.5/SDXL, models finetuned with DPO and models finetuned with the proposed Adapter-DPO. Our method enjoys generally better image quality.

\section{Additional Results for Portrait images}

Additionally, we train SD1.5 with self-collected portrait images using both Diffusion-DPO and Adaptive-DPO, the above pictures in Fig.\ref{fig:supp-ablation} shows us the results. Our methods achieves fewer deformities, clearer facial features, and high-quality portrait results.

\newpage

{
    \small
    \bibliographystyle{ieeenat_fullname}
    \bibliography{main}
}


\end{document}